\documentclass{article}
\usepackage{iclr2027_conference,times}
\iclrfinalcopy

\usepackage{mathtools,amssymb,latexsym,amsfonts}
\usepackage{graphicx}
\usepackage{xcolor}
\usepackage{enumitem}
\usepackage{booktabs,array,multirow}
\usepackage{subcaption}
\usepackage{placeins}
\usepackage{xspace}
\usepackage{algorithm}
\usepackage[noend]{algpseudocode}
\usepackage{hyperref}
\usepackage{xurl}
\usepackage[capitalise]{cleveref}
\hypersetup{
  hidelinks,
  pdftitle={\textsc{Fiona}: Accelerating FHE Inference with Packing-Aware Ternary Weights},
  pdfauthor={Yiteng PENG, Zhibo LIU, Dongwei XIAO, Shuai WANG}
}
\newcommand{\tool}{\textsc{Fiona}\xspace}
\newcommand{\F}{Fig.}
\newcommand{\E}{Eq.}
\newcommand{\T}{Table}
\renewcommand{\S}{Sec.}
\newcommand{\A}{Alg.}

\newcommand{\PMult}{\ensuremath{\mathsf{PMult}}\xspace}
\newcommand{\CMult}{\ensuremath{\mathsf{CMult}}\xspace}
\newcommand{\Add}{\ensuremath{\mathsf{Add}}\xspace}
\newcommand{\Sub}{\ensuremath{\mathsf{Sub}}\xspace}
\newcommand{\Rot}{\ensuremath{\mathsf{Rot}}\xspace}

\newcommand{\Plan}{\ensuremath{\Pi}\xspace}
\newcommand{\calib}{\mathcal{D}_{\mathrm{cal}}}
\newcommand{\groups}{\mathcal{G}}

\newcommand{\parh}[1]{\noindent\textbf{#1}}

\newcommand{\stage}[1]{\textbf{\textcircled{\normalsize #1}}}
\newcommand{\Description}[1]{}
 
\title{\tool: Accelerating FHE Inference with Packing-Aware Ternary Weights}
\author{Yiteng PENG$^1$ \quad Zhibo LIU$^2$ \quad Dongwei XIAO$^1$ \quad Shuai WANG$^1$\\
{\normalfont $^1$The Hong Kong University of Science and Technology}\\
{\normalfont $^2$State Key Laboratory of Novel Software Technology, Nanjing University}\\
{\normalfont\texttt{\{ypengbp,dxiaoad,shuaiw\}@cse.ust.hk}}\\
{\normalfont\texttt{zhiboliu@nju.edu.cn}}}

\begin{document}
\maketitle
\lhead{Preprint}

\begin{abstract}
Fully homomorphic encryption (FHE) enables neural network inference directly on
encrypted inputs, but it remains orders of magnitude slower than plaintext
inference.
Applying the server's plaintext weights to encrypted activations involves
plaintext--ciphertext multiplications (PMult) and accounts for more than
half of inference time in recent systems.
Ternary quantization can replace these multiplications with additions and
subtractions, but the savings rarely materialize under packed execution.
A single PMult applies a weight group fixed by the packing layout and can
be avoided only when all its weights share the same ternary value. Ternarizing all groups, however, largely degrades accuracy.

We present \tool, an offline optimizer that selectively ternarizes
weights within a given packing layout based on the estimated effect
of ternary conversion on the model's performance.
\tool encourages a shared ternary value within each weight group
and retains full-precision weights for sensitive groups, so
ternarized and full-precision paths coexist within a layer.
It then compiles these hybrid operators exactly, applying common
scaling factors once to accumulated inputs and reusing sums across outputs.
Weight ternarization can also narrow the input ranges of downstream
polynomials.
\tool fits lower-degree replacements under a cumulative accuracy budget,
reducing multiplicative depth and bootstrapping.
On VGG11, ViT, and BERT, \tool reduces PMult operations by
53.4--79.5\% and accelerates end-to-end encrypted inference by
$2.38{\times}$, $1.68{\times}$, and $1.84{\times}$, respectively,
with less than 1\% accuracy loss across all three models.
 \end{abstract}

\section{Introduction}
\label{sec:intro}

Cloud-hosted deep learning (DL) services support applications involving
sensitive data, such as medical images and private text. Fully homomorphic
encryption (FHE) enables these services by letting the server compute directly
on encrypted inputs~\citep{gentry2009fully,cheon2017ckks,zhang2025nexus}. The
client encrypts its input and sends the ciphertext to the server, which
evaluates the model on it using weights that it holds in plaintext and returns
an encrypted prediction that only the client can
decrypt~\citep{gilad2016cryptonets}. 
Despite these benefits, the computational cost of FHE remains
a major barrier to deployment. To reduce computational cost, existing FHE-DL
systems optimize ciphertext packing and execution flows through convolution
packing~\citep{ebel2025orion}, column and diagonal packing~\citep{zhang2026moai},
or column-wise input encoding~\citep{he2025ensi}; other approaches approximate
nonlinear activation operators with
polynomials~\citep{ao2024autofhe,xie2026uldnet}

However, a major source of this cost is linear computation with plaintext
weights, which involves plaintext--ciphertext multiplication (\PMult). A recent
GPU profile of FHE-protected LLaMA-3-8B attributes up to 68\% of its runtime to
such plaintext--ciphertext matrix multiplication~\citep{zhang2026moai}. Unlike
the activations, the weights are plaintext and fixed before any query arrives,
so their representation can be optimized offline, and every encrypted operation
removed in this way is saved on each later inference. A few approaches do
optimize weight computation, yet their designs are tailored to particular
encodings or model architectures, through sub-block pruning for CNNs in
SpENCNN~\citep{ran2023spencnn} and block-circulant weights with a matching encoding
in PrivCirNet~\citep{xu2024privcirnet}, so their savings do not
automatically carry over to different packing layouts.  We therefore study how
to reduce weight multiplications by changing weight values under a backend's existing packing layout.

A ternary weight of $+1$, $-1$, or $0$ needs no multiplication in a weighted
sum, because the
server can add the encrypted input to the output, subtract it, or skip it. 
In our measurement, a weighted sum of 64 ciphertexts is
$14.8\times$ faster than the same sum computed with \PMult
(\T~\ref{tab:motivation-kernel}). 
Ternary quantization~\citep{wang2023bitnet,ma2024bitnetb158}, which restricts
each weight to one of these three values times a scaling factor, is
therefore a natural starting point for reducing weight multiplications.

However, ternary weights alone do not remove \PMult operations, because a
backend does not always apply weights one at a time. CKKS~\citep{cheon2017ckks}, the FHE
scheme that most encrypted inference systems adopt, packs a vector of values
into each ciphertext, and one \PMult applies a vector of weights to it.
The backend's \emph{packing layout} determines which weights share one such
vector, and we call each such set of weights an \emph{execution group}. An
addition or a subtraction also acts on a whole ciphertext, so the server can
avoid the \PMult only when all weights in the group have the same ternary value.
We call such a group \emph{pure}. A group that holds different values, such as
$+1$ and $0$, still requires the \PMult.

Standard ternary quantization does not account for this structure, and it is
ineffective in two respects. Scalar ternary quantization of ResNet-8
leaves only 4.06\% of the execution groups pure, so the remaining 95.94\%
still require a full \PMult, and it also reduces CIFAR-10 accuracy from 87.53\% to
82.43\% (\S~\ref{sec:motivation}). These observations yield two requirements
that are in tension. \textit{(1) Group-level uniformity.} A reduction in cost
is obtained per execution group rather than per weight, so the ternary
weights of a converted group must share a single value. \textit{(2)
Selective conversion.} Conversion reduces accuracy, so it must be restricted
to the groups whose conversion the task can tolerate, and the remaining groups
must retain their original full-precision weights.

We present \tool, an offline optimizer that makes this decision group by group
and then compiles the result. \tool takes a network, its polynomial
approximations, and the backend's packing layout, and produces an execution plan
that the server replays on every query. It first trains the weights to encourage
each group to adopt a shared ternary value, applying stronger pressure to
groups that are less sensitive to ternary conversion and leaving the most
sensitive groups unconverted. The measure it uses is a group's \emph{task
sensitivity}, an estimate of how much the model's loss would rise if that
group were converted. Every group therefore takes one of two
\emph{routes}, a cheap signed route or an ordinary multiplication route, and
both coexist within a layer. In addition, weight ternarization can narrow
the input ranges of the polynomials used for nonlinear operators. \tool fits
lower-degree replacements and selects those that reduce
multiplicative depth, further accelerating encrypted inference.

In summary, this paper makes the following contributions:
\begin{itemize}[leftmargin=*,itemsep=2pt,parsep=0pt,topsep=3pt]
    \item A hybrid weight representation that aligns ternary
    structure with packed execution. Task sensitivity guides ternarizations in fixed packing groups and retention of full-precision weights
    to balance encrypted cost and accuracy.

    \item An exact compilation step that applies one reconstruction factor per
    signed sum and shares recurring sums across outputs, followed by
    lower-degree polynomial refitting on the routed model under a
    cumulative accuracy budget.
    
    \item An implementation evaluated on CNN and Transformer models for image
    and text classification. On VGG11, ViT, and BERT, \tool removes 53.4\%
    to 79.5\% of \PMult operations and 37.0\% to 57.1\% of ciphertext
    refreshes, making encrypted inference $2.38\times$, $1.68\times$, and
    $1.84\times$ faster than full-precision baselines, with less than 1\% accuracy loss.

\end{itemize}
 \section{Preliminaries}
\label{sec:background}

\subsection{Encrypted DL Inference with FHE}
\label{sec:background-setting}

\parh{Setting.}
We consider inference between a client that holds a private input and a server
that hosts the model. FHE lets the server
compute on ciphertexts without access to the secret key, so the encrypted
output decrypts to the result of the same computation on the plaintext
input~\citep{cheon2017ckks}. The client encrypts its input, sends it to the
server, and decrypts the returned prediction. The server holds the model
weights in plaintext, and we call them \emph{plaintext weights}; the input, the
intermediate activations, and the prediction are encrypted.

\parh{Threat Model.}
We assume an honest-but-curious server that follows the prescribed protocol but
may try to infer private information from the data it processes. The semantic
security of the underlying FHE scheme protects the client's input values and
prediction from the server. Input shapes and padding masks are public
metadata.

\subsection{Packed CKKS Execution}
\label{sec:background-ckks}

\parh{Ciphertexts and Slots.}
Among modern FHE
schemes~\citep{fan2012somewhat,brakerski2014leveled,chillotti2020tfhe}, we focus
on CKKS~\citep{cheon2017ckks}, which supports approximate arithmetic on vectors
of real numbers and is the scheme adopted by most recent work on encrypted
neural network
inference~\citep{ao2024autofhe,ebel2025orion,zhang2025nexus,zhang2026moai}.
One CKKS ciphertext holds a vector of values rather than a single number,
each in its own \emph{slot}, and one homomorphic operation processes all
slots in parallel.

\parh{Homomorphic Operations.}
A computation on CKKS ciphertexts can be represented as an \emph{arithmetic circuit}, that is,
a fixed sequence of additions and multiplications which always involves five
basic operations.
Addition and subtraction (\Add, \Sub) combine two ciphertexts slot by slot, and
a rotation (\Rot) cyclically shifts the slots so that values in different slots
can be added together. The other two operations are multiplications, and
they differ in their second operand. A \PMult multiplies a ciphertext slot by
slot with a plaintext vector, whereas a ciphertext--ciphertext multiplication
(\CMult) multiplies two ciphertexts. The weights are plaintext and the
activations are encrypted, so a convolution or a matrix multiplication applies
its weights with \PMult and sums the products with \Rot and \Add. \CMult is
needed only where two encrypted values are multiplied, for example, when a
polynomial activation squares its input or when attention multiplies the query
and key matrices. In our measurement, computing a weighted sum with \Add and \Sub
takes less than one tenth of the time with \PMult
(\T~\ref{tab:motivation-kernel}), removing most of the computation cost.

\parh{Levels and Bootstrapping.}
CKKS attaches a scaling factor to each ciphertext, and a
multiplication multiplies the factors of its two operands. The evaluator
restores the scale by \emph{rescaling}, which consumes one \emph{level} from a
finite modulus chain fixed at key generation,
so a ciphertext supports only a limited number of successive
multiplications. The number of levels a circuit needs is its
\emph{multiplicative depth}, that is, the longest chain of dependent
multiplications. When a circuit
needs more levels than remain available, \emph{bootstrapping} refreshes the
ciphertext at a substantially higher cost than ordinary homomorphic
operations~\citep{cheon2018bootstrapping, cheon2019full, ao2024autofhe,ebel2025orion}. Fewer multiplications reduce arithmetic work, while lower
multiplicative depth can reduce bootstrapping operations.

\parh{Execution Plan.}
The operations, plaintext operands, and rescaling and bootstrapping
schedule are independent of encrypted input values and fixed before any
query arrives. We call this fixed program an \emph{execution plan} $\Plan$. The model provider builds it offline from its training or
calibration data, and the server replays it on every query.

\subsection{Plaintext Weights under a Packing Layout}
\label{sec:background-layout}

\parh{Packed Evaluation of a Linear Layer.}
To evaluate a convolution or a matrix multiplication, a backend maps tensor
elements to ciphertext slots and encodes the layer's weights
as plaintext operands. It applies these operands with \PMult and combines the
results with \Rot and
\Add~\citep{dathathri2020eva,ran2023spencnn,zhang2025nexus}. Each \PMult acts
on the entire packed operand, regardless of the values of the weights
inside it.

\parh{Packing Layout and Execution Groups.}
We call this backend-specific mapping a \emph{packing layout} $\Gamma$. 
For layer $l$, the layout $\Gamma_l$ specifies which weights share one
plaintext operand. We call each such group an \emph{execution group} and denote
the layer's groups by $\mathcal G_l$.
For example, under the convolution layout we evaluate, a group contains the
weights on one diagonal of an output-channel block at a fixed
kernel position (\S~\ref{sec:implementation}). 
The layout determines the granularity at which plaintext weights are applied, so whether a weight
multiplication can be simplified depends on the values across the entire
execution group.

\parh{Ternary Weight Quantization.}
Ternary quantization replaces each weight with a value in $\{-1,0,+1\}$ scaled
by a per-output-channel factor, which we call the \emph{reconstruction factor},
so that, in a weighted sum, a product can become an addition, a subtraction, or
a skipped term~\citep{wang2023bitnet,ma2024bitnetb158}. Since rounding is not
differentiable, such networks are trained with quantization-aware training~\citep{jacob2018quantization} and a
straight-through estimator~\citep{bengio2013estimating} for the rounding step (\S~\ref{sec:method-shaping}).

\subsection{Polynomial Operators}
\label{sec:background-poly}

\parh{FHE-friendly Polynomial Networks.}
CKKS evaluates only additions and multiplications, so nonlinear
operators, such as ReLU, GELU, Softmax, and the reciprocal square root in
LayerNorm, cannot be evaluated directly. Replacing them with a
polynomial approximation, or with an alternative using only supported
arithmetic, and training the network to tolerate the replacement, is standard
practice in non-interactive FHE
inference~\citep{ao2024autofhe,xie2026uldnet,zimerman2024powersoftmax,nam2025slothe}.
We call the resulting graph an \emph{FHE-friendly polynomial
network}. \tool takes such a network, with its supplied activation and
normalization polynomials, as input and optimizes its weights. 

\parh{Polynomial Cost and Approximation Range.}
A polynomial is itself an arithmetic circuit, so its degree and evaluation
strategy determine the number of multiplications it performs and the depth it
adds. Lower-degree approximations can reduce both and allow more layers to be evaluated
within the available levels, but lowering the degree too far increases
approximation error and degrades accuracy. Each approximation is fitted over a
specified input interval, and at the same error tolerance a narrower interval
allows a lower degree. In a network, the interval must cover the values that
the operator receives. The preceding layers compute these values from their
weights, so changing the weights can narrow the interval and allow a lower
degree (\S~\ref{subsec:ob4}).
 \section{Motivation}
\label{sec:motivation}

A ternary weight needs no multiplication, but in packed FHE one \PMult
applies all weights of an execution group at once
(\S~\ref{sec:background-layout}), so the server can drop the \PMult only when
every weight in the group takes the same ternary value. Training weights
toward this condition, in turn, can reduce model accuracy. 
We quantify the cost of plaintext weights (Observation~1),
the efficiency and accuracy limits of scalar ternarization
(Observation~2), and its potential to reduce polynomial cost
(Observation~3), and then state what they imply
for the design of \tool.

\subsection{Observations}
\label{sec:mot-observations}

\begin{figure}[t]
\centering
\includegraphics[width=\columnwidth]{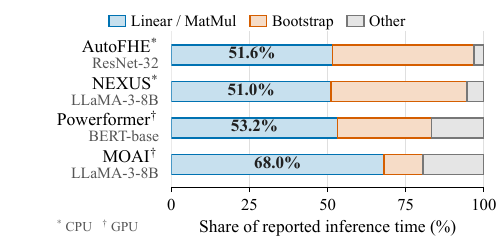}
\Description{Time breakdown across four published inference configurations,
showing computation with plaintext weights, bootstrapping, and other execution time.}
\caption{FHE inference time breakdowns reported in recent work.}
\label{fig:motivation-public-weight}
\end{figure}

\parh{Observation 1. Plaintext weights remain a major cost.}
\F~\ref{fig:motivation-public-weight} breaks down four published FHE
inference profiles, covering one CNN and three Transformer models on CPUs and
GPUs~\citep{ao2024autofhe,zhang2025nexus,park2025powerformer,zhang2026moai},
into computation with plaintext weights, bootstrapping, and other operations.
Computation with plaintext weights takes 51.0--68.0\% of the reported
inference time, more than half in every profile. On AutoFHE's ResNet-32, for
example, it takes 51.6\%, compared with 45.3\% for bootstrapping.

\parh{Observation 2. Ternary quantization alone rarely removes a \PMult.}\label{subsec:ob3}
We call an execution group \emph{pure} when all its weights have the same
ternary value, and \emph{mixed} otherwise. For a pure group, the server
replaces the \PMult by an \Add, a \Sub, or a skip of the input ciphertext. A
mixed group still needs the \PMult~\citep{he2025ensi}; splitting it into a
positive and a negative part instead requires selector masks and additional
operations.

\T~\ref{tab:motivation-kernel} compares these paths on a weighted sum of 64
input ciphertexts. With full-precision weights and with mixed ternary groups,
the sum takes nearly the same time, 14.16 and 14.15\,ms, because both cases use
\PMult. With pure ternary groups, the sum uses only \Add and \Sub, and it takes
0.96\,ms, $14.8\times$ faster. Splitting mixed groups with sign selectors is
slower than the \PMult, at 26.51\,ms. Whether a group can avoid the \PMult
is therefore decided by all of its weights together, not by any single weight.

\begin{table}[t]
\centering
\caption{Kernel latency for a weighted sum of 64 packed inputs
($N=8192$, $scale=2^{40}$). Values are medians of five runs after one warmup.}
\label{tab:motivation-kernel}
\small
\setlength{\tabcolsep}{2.5pt}
\renewcommand{\arraystretch}{0.95}
\begin{tabular}{lcc}
\toprule
Case & Path & Latency \\
\midrule
Full precision & \PMult & 14.16 ms \\
mixed ternary & \PMult & 14.15 ms \\
pure ternary & \Add/\Sub & 0.96 ms \\
Selector split & Mask+\Add/\Sub & 26.51 ms \\
\bottomrule
\end{tabular}
\end{table}

Scalar ternary quantization, which rounds each weight on its own, produces
few pure groups. We trained a ResNet-8 on CIFAR-10 with scalar ternary
quantization-aware training (QAT~\citep{jacob2018quantization}) and split the
flattened weights of each output channel into consecutive groups of eight. Of
the 9,664 complete groups, 4.06\% are pure and 95.94\% are mixed. Since a
weight's ternary value is the rounding of its trained value, a group becomes
pure only if training moves all of its weights to the same value, and the scalar
QAT objective does not encourage this.

Additionally, on CIFAR-10, a full-precision ResNet-8 reaches 87.53\% accuracy, and the
same model trained with scalar ternary QAT on all weights reaches 82.43\%. A
partial variant of the same QAT, which ternarizes a fixed random subset of
about half the weights and keeps the other weights at full precision, reaches
84.31\% on average over three seeds. Partial conversion thus retains more
accuracy than full conversion in this comparison.

\parh{Observation 3. Ternarization allows lower-degree polynomials.}\label{subsec:ob4}
Ternarization also changes the outputs of a linear layer, which are the
inputs of the polynomial that follows it. Consider one output, $w^\top x$,
with $n$ weights $w\in\mathbb R^n$ and input vector $x\in\mathbb R^n$. Ternary
quantization replaces $w$ with $\gamma q$, where $q\in\{-1,0,+1\}^n$ holds the
ternary values and $\gamma=\frac{1}{n}\sum_{i=1}^{n}|w_i|$, the mean absolute
weight, is the reconstruction factor (\S~\ref{sec:background-layout}). If the
entries of $x$ are uncorrelated and have the same variance $\sigma_x^2$, that
is, $x$ has covariance matrix $\sigma_x^2 I$ with $I$ the $n\times n$ identity
matrix, then
\begin{equation}
\begin{aligned}
\mathrm{Var}((\gamma q)^\top x)
  &=\sigma_x^2\gamma^2\|q\|_2^2
  \leq\sigma_x^2\|w\|_2^2=\mathrm{Var}(w^\top x).
\end{aligned}
\label{eq:ternary-variance}
\end{equation}
The equalities use $\mathrm{Var}(a^\top x)=\sigma_x^2\|a\|_2^2$ for any
fixed $a\in\mathbb R^n$. The inequality holds because $\|q\|_2^2\le n$, as
every entry of $q$ is $-1$, $0$, or $+1$, and because
$n\gamma^2=\frac{1}{n}(\sum_{i}|w_i|)^2\le\sum_{i}w_i^2$ by the
Cauchy--Schwarz inequality~\citep{steele2004cauchy}. The variance of the
ternary output is thus at most that of the full-precision output, so the
polynomial's input interval can be narrower, which allows a lower degree at
the same error (\S~\ref{sec:background-poly}).

\begin{figure}[t]
\centering
\includegraphics[width=\columnwidth]{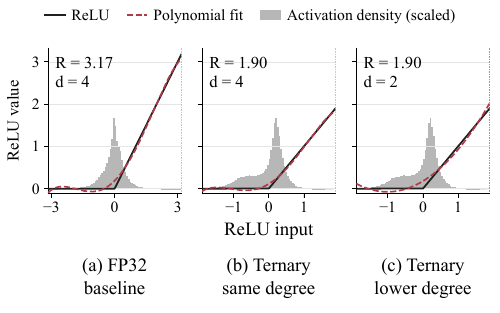}
\Description{Input distributions and ReLU polynomial fits at the first ReLU
in the first residual block of ResNet-8. The ternary model has a smaller
99.9th-percentile input radius and lower approximation error at this site.}
\caption{Input distribution and polynomial fits at the first ReLU
in the first residual block
of full-precision and scalar-ternary ResNet-8. The radius $R$ is the 99.9th
percentile of absolute sampled inputs. Polynomials approximate ReLU directly
by least squares on $[-R,R]$.}
\label{fig:motivation-corridor}
\end{figure}

We checked this effect at the first ReLU in the first residual block of
ResNet-8, comparing the full-precision and the scalar ternary model
(\F~\ref{fig:motivation-corridor}). The input radius, the 99.9th percentile
of the absolute inputs, decreases from 3.17 to 1.90. For a degree-4
polynomial fitted to ReLU on the corresponding interval, the root mean squared
error (RMSE) falls from 0.1169 to
0.0454, and a degree-2 polynomial on the ternary model, with an RMSE of
0.0861, is still more accurate than degree 4 on the full-precision model. At
this site, ternarization therefore allows a lower degree at a lower error.

\begin{table}[t]
\centering
\caption{Model-side optimization across approaches.}
\label{tab:motivation-positioning}
\small
\setlength{\tabcolsep}{2.5pt}
\renewcommand{\arraystretch}{0.95}
\begin{tabular*}{0.9\columnwidth}{@{\extracolsep{\fill}}lccc@{}}
\toprule
\multicolumn{1}{l}{\multirow{2}{*}{Work}} & Weight & Relation to & Low-degree \\
 & Str. Opt. & packing & design \\
\midrule
SpENCNN~\citep{ran2023spencnn} & $\checkmark$ & Co-designed & $\times$ \\
PrivCirNet~\citep{xu2024privcirnet} & $\checkmark$ & Co-designed & $\times$ \\
ENSI~\citep{he2025ensi} & $\times$ & Co-designed & $\times$ \\
AutoFHE~\citep{ao2024autofhe} & $\times$ & --- & $\checkmark$ \\
ULD-Net~\citep{xie2026uldnet} & $\times$ & --- & $\checkmark$ \\
\textbf{\tool} & $\checkmark$ & Given layout & $\checkmark$ \\
\bottomrule
\end{tabular*}
\par\smallskip
\footnotesize\raggedright
Weight Str. Opt.: methods that optimize weight structure to reduce packed linear cost;
Packing: relation between weight representation and encoding (---: no joint design);
Low-degree design: dedicated methods for low-degree polynomial networks
or degree reduction.
$\checkmark$: covered; $\times$: not covered.\par
\end{table}

\subsection{Implications for Design}
\label{sec:mot-implications}

These observations lead to three design decisions. Observation~1 shows
that plaintext weights are the largest cost, and Observation~2 that the unit
of saving is the execution group, so \tool ternarizes weights under the
backend's fixed packing layout and trains each group toward one shared ternary
value (\S~\ref{sec:method-shaping}). Observation~2 additionally shows that converting all
groups costs accuracy, so \tool converts only the groups whose conversion the
task tolerates, judged by each group's \emph{task sensitivity}, the estimated
rise in task loss from converting it, and keeps the original, or \emph{raw},
weights elsewhere. A layer then mixes pure groups on the \Add/\Sub path with
raw groups on the \PMult path, and \tool compiles both paths exactly
(\S~\ref{sec:method-reuse}). Observation~3 shows that ternarization changes
the polynomials' inputs, so \tool selects polynomial degrees after the weights
are fixed, fitting each polynomial on the converted model's input range under
an accuracy budget (\S~\ref{sec:method-corridor}).

\subsection{Positioning of Our Approach}
\label{sec:mot-positioning}

Existing FHE-DL systems mainly improve packing and
dataflow~\citep{ebel2025orion,zhang2026moai} or reduce polynomial
computation through degree selection and low-degree network
training~\citep{ao2024autofhe,xie2026uldnet}.
Among approaches targeting weight computation, SpENCNN co-designs sub-block pruning and packing~\citep{ran2023spencnn},
while PrivCirNet uses block-circulant weights with a matching
encoding~\citep{xu2024privcirnet}. ENSI instead executes
already-ternarized linear layers using column-wise
packing~\citep{he2025ensi}.
\T~\ref{tab:motivation-positioning} summarizes these optimization
choices.

\tool optimizes weights within the execution groups of a given
packing layout. Guided by task sensitivity, it encourages a
common ternary state within each group and retains raw weights
in selected sensitive groups. The resulting hybrid
operators reduce multiplications for nonzero weight contributions
while retaining full precision where needed. Exact compilation
further reduces repeated reconstruction and accumulation work.
With weights fixed, \tool fits and selects lower-degree
polynomials using the resulting input distributions, exploiting
the approximation opportunities created by weight optimization.
 \section{Design of \tool}
\label{sec:method}

\begin{figure}[t]
\centering
\includegraphics[width=\textwidth]{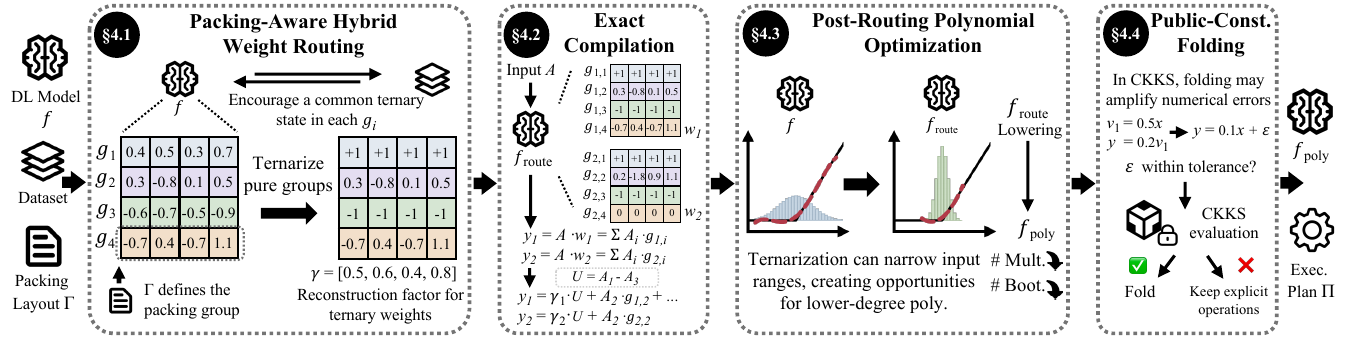}
\Description{Four offline stages: weight routing, linear compilation,
polynomial optimization, and public-constant folding. The packing layout
guides the first two stages. The final plan feeds a separate CKKS executor.}
\caption{Overview of \tool.}
\label{fig:pipeline-overview}
\end{figure}

Given a polynomial network with plaintext weights and a fixed packing layout
$\Gamma$, \tool uses the model provider's data for training and calibration
to construct a CKKS execution plan $\Plan$ offline.
\F~\ref{fig:pipeline-overview} shows the four stages.

\parh{\stage{1} Packing-Aware Hybrid Weight Routing.}
\tool optimizes the weights with a loss that pushes each execution group
toward one shared ternary value, weighting the loss by the group's task
sensitivity, and keeps the raw weights of the most sensitive groups.
The output model fixes the ternary/full-precision routes and reconstruction
factors. 
These partially quantized weights, together with the supplied activation and
normalization polynomials, define the routed model $f_{\mathrm{route}}$.

\parh{\stage{2} Exact Compilation of Routed Linear Operators.} With routes fixed, 
\tool~rewrites each linear layer in three ways. It applies the reconstruction
factors once to a signed sum of input ciphertexts rather than once per input,
computes a signed sum once when several outputs share it, and adds an output's
signed and raw results before the rescaling that follows a \PMult.
Each rewrite is an algebraic identity, so the layer computes the same
function with fewer \PMult, \Add/\Sub, and rescaling operations.

\parh{\stage{3} Post-routing Polynomial Optimization.} Then, for the routed
model $f_{\mathrm{route}}$, \tool fits lower-degree candidates 
for the supplied polynomials on a calibration dataset and accepts
the replacements that reduce the multiplicative depth within a cumulative
accuracy budget, producing
$f_{\mathrm{poly}}$.

\parh{\stage{4} Public-Constant Folding and Plan Export.}
\tool folds public constants into adjacent operators, 
keeping the unfolded form wherever a numerically sensitive fold fails an
error check against it. It then removes unused operations and exports the final
execution plan $\Plan$.

\subsection{Packing-Aware Hybrid Weight Routing}
\label{sec:method-shaping}

As \S~\ref{sec:mot-implications} concluded, \tool must make execution groups
pure under the fixed packing layout $\Gamma$, and it must leave the groups whose
conversion the task cannot tolerate at their raw weights. This stage does both
during training, guided by task sensitivity. We describe the weight
representation, the sensitivity estimate, the training objective, and the
protection of sensitive groups in turn.

\parh{Packing-Aware Weight Representation.}
During ternary quantization, \tool computes a public reconstruction
factor $\gamma_o$ for each output channel $o$ as the mean absolute
value of its weights.
Let $W_{o,i}$ be the $i$th of the $n$ weights feeding channel $o$.
The reconstruction factor and ternary candidates are
\begin{equation}
\begin{aligned}
\gamma_o&=\frac{1}{n}\sum_{i=1}^{n}|W_{o,i}|,\\
q_{o,i}&=\operatorname{clip}_{[-1,1]}
\left(\operatorname{round}\frac{W_{o,i}}{\gamma_o}\right),\qquad
\widehat W_{o,i}=\gamma_o q_{o,i}.
\end{aligned}
\label{eq:ternary-candidate}
\end{equation}
Rounding to the nearest integer and clipping to $[-1,1]$ yields
ternary values in $\{-1,0,+1\}$. For $\gamma_o=0$, we set $q_{o,i}=0$.
The layout groups these values and candidate weights into $q_g$ and
$\widehat W_g$, respectively.

A group $g$ is \emph{pure} when all its ternary values are identical, i.e.,
$q_g=h_g\mathbf 1$ for $h_g\in\{-1,0,+1\}$, where $\mathbf 1$ is the
all-ones vector, and \emph{mixed} otherwise. 
A pure group can take the \emph{signed route}. The server adds the input
ciphertext when $h_g=+1$, subtracts it when $h_g=-1$, and skips it when $h_g=0$,
and it applies the reconstruction factors afterwards.
For example, the groups $(1,1,1,1)$ and
$(-1,-1,-1,-1)$ replace their multiplications by one addition or one
subtraction, whereas $(1,0,1,-1)$ still requires a vector \PMult.
Every other group takes the \emph{raw route}, one \PMult by the group's raw weights. Since
a group may span several output channels, its weights may have different
reconstruction factors; \S~\ref{sec:method-reuse} describes how the compiler
applies them as one packed operand.

Task Sensitivity.
Both decisions of this stage depend on how much the task loss would rise
if a group were converted. \tool estimates this rise for group $g$ from the
candidate weight change $\Delta_g=\widehat W_g-W_g$ as
\begin{equation}
F_g=\sum_{i\in g}D_i\Delta_{g,i}^{2},
\label{eq:group-fisher-proxy}
\end{equation}
where $D_i$ estimates the curvature of the task loss with respect to weight
$i$,
\begin{equation}
D_i=\mathbb E_{\mathrm{data}}\!\left[
\sum_p\left(x_p^{(i)}
\frac{\partial\ell_{\mathrm{task}}}{\partial z_p^{(i)}}\right)^2
\right].
\label{eq:weight-fisher-proxy}
\end{equation}
Here $\ell_{\mathrm{task}}$ is the task loss of one training sample,
$\mathbb E_{\mathrm{data}}$ averages over the training samples, and $p$ indexes
the uses of weight $i$ within one sample, such as the spatial positions of a
convolution or the tokens of a linear layer. At use $p$, weight $i$
multiplies the input value $x_p^{(i)}$, and the product enters the layer output
$z_p^{(i)}$, so the summand is the squared contribution of that use to the
gradient of the loss with respect to weight $i$. Thus $D_i$ is a per-use form
of the empirical Fisher diagonal, a standard estimate of the loss curvature
under parameter changes~\citep{kirkpatrick2017overcoming}, and $F_g$ is the
resulting second-order estimate, up to a constant factor, of the loss increase
caused by $\Delta_g$~\citep{lecun1989optimal}. We call $F_g$ the \emph{task
sensitivity} of group $g$ (\S~\ref{sec:intro}).

\parh{Sensitivity-Weighted Homogeneity Loss.}
To make groups pure, \tool adds to the task loss a \emph{homogeneity loss}
that rewards agreement among the ternary values within a group. The rounding in
\E~\ref{eq:ternary-candidate} has zero gradient almost everywhere, so the loss
uses a \emph{soft assignment} of each weight to the three states. Let
$u_i=W_{o,i}/\gamma_o$ be weight $i$ divided by its reconstruction factor, so
that the rounding rule maps $u_i$ between $-1/2$ and $1/2$ to $0$,
and smaller and larger values to $-1$ and $+1$. \tool computes
$s_i(0)=\sigma(\kappa(1/2-|u_i|))$ and $s_i(\pm1)=\sigma(\kappa(\pm u_i-1/2))$,
where $\sigma$ is the sigmoid function and the fixed slope $\kappa$
controls how sharply the scores change around these boundaries. The soft
assignment of weight $i$ to state $c$ is the normalized score
$p_i(c)=s_i(c)/\sum_{c'\in\{-1,0,+1\}}s_i(c')$. It is differentiable in the
weight, and it approaches the rounding rule as $\kappa$ grows.

The homogeneity loss of group $g$ is
\begin{equation}
\mathcal L_{\mathrm{hom},g}
=-\frac{1}{|g|}\log\sum_{c\in\{-1,0,+1\}}\prod_{i\in g}p_i(c).
\label{eq:sensitivity-shaping}
\end{equation}
If each weight were assigned a state independently according to $p_i$, the
sum would be the probability that all weights of $g$ receive one common state,
and the loss is the negative logarithm of this probability per
weight. Minimizing it therefore moves the group toward one shared ternary
state.

The training objective is
\begin{equation}
\mathcal L=\mathcal L_{\mathrm{task}}
+\frac{\lambda_{\mathrm{group}}}{|\groups|}
\sum_{g\in\groups}a_g\mathcal L_{\mathrm{hom},g}.
\label{eq:group-training-objective}
\end{equation}
Here $\mathcal L_{\mathrm{task}}$ is the task loss, $\lambda_{\mathrm{group}}$
sets the overall strength of the regularization, and $a_g$ weights group $g$ by
its task sensitivity $F_g$ so that less sensitive mixed groups are pushed harder.
After each epoch, \tool normalizes the scores $F_g$ of the groups that were
mixed in that epoch by subtracting their mean
and dividing by their standard deviation across layers. It maps the normalized scores
through a decreasing sigmoid, so a mixed group with lower sensitivity receives
a larger $a_g$. Pure groups receive $a_g=1$.

\parh{Protecting Sensitive Pure Groups.}
The homogeneity loss makes groups pure, but a pure group may still be one
whose conversion the task cannot tolerate. \tool therefore \emph{protects} the
most sensitive pure groups by keeping their raw weights, so that a protected
group takes the raw route although it is pure. Protection is decided once per
epoch from $F_g$ averaged over the preceding epoch, within each \emph{packing
pool}, the set of groups across layers that share the same grouping rule and
group size under $\Gamma$ (\T~\ref{tab:implementation-layouts}). With
$\rho_{\mathrm P,\max}$ the maximum protection ratio, \tool limits the fraction
of groups on the signed route in each pool to $1-\rho_{\mathrm P,\max}$. If the
fraction of pure groups exceeds this limit, it protects the pure groups with
the highest $F_g$ until the limit is met; otherwise it protects none. For
example, with $\rho_{\mathrm P,\max}=0.2$ and 90 pure groups among 100, the 10
most sensitive pure groups are protected.

Let $\mathcal H$ denote the protected groups and $\mathcal P_\Gamma$ the
pure groups under $\Gamma$.
The weights used in the forward pass are
\begin{equation}
\widetilde W_g=
\begin{cases}
\widehat W_g, & g\in\mathcal P_\Gamma\setminus\mathcal H,\\
W_g, & \text{otherwise}.
\end{cases}
\label{eq:hybrid-route}
\end{equation}
An unprotected pure group thus uses its ternary candidate, and every other
group, mixed or protected, uses its raw weights. All groups remain trainable
and contribute to the homogeneity loss, including protected groups.

\parh{Training Schedule and Output.}
Training uses $\widetilde W_g$ in the forward pass and a straight-through
estimator for the rounding in \E~\ref{eq:ternary-candidate}, so that gradients
reach the raw weights of every group (\S~\ref{sec:background-layout}). Group
purity is recomputed on each forward pass, whereas the protected set
$\mathcal H$ and the sensitivity scores used for loss weighting are updated once per epoch and stay fixed
within it. After training, \tool freezes the weights, the routes, and the
reconstruction factors. The result, with the supplied polynomials, is the
routed model $f_{\mathrm{route}}$.

\subsection{Exact Compilation of Routed Linear Operators}
\label{sec:method-reuse}

With weights and routes fixed, a linear layer of $f_{\mathrm{route}}$
computes each packed output from \emph{signed terms} (i.e., input ciphertexts
that pure groups add or subtract) and \emph{raw terms} (i.e., input ciphertexts
that raw groups multiply). Evaluated term by term, every signed term still needs
one \PMult by the reconstruction factors, so the signed route would remove no
\PMult. This stage therefore rewrites the layer in three ways. It applies the
reconstruction factors once to an output's \emph{signed sum}, the sum of its
signed terms; it computes a signed sum once when several outputs share it; and
it adds an output's signed and raw results before the rescaling that follows a
\PMult. The rewrites preserve the layer's function over real arithmetic and
reduce CKKS operations.

Consider the example in \F~\ref{fig:pipeline-overview}. Two packed outputs
$y_1$ and $y_2$ are computed from four input ciphertexts $A_1,\dots,A_4$,
prepared according to the packing layout. In output $y_j$, $g_{j,i}$ denotes
the group applied to $A_i$ and also its vector of raw weights, which multiplies
$A_i$ slot by slot, and $\gamma_j$ is the vector of reconstruction factors of
the output channels packed in $y_j$. 
The groups $g_{1,1}$ and $g_{2,1}$ are pure with $h_g=+1$, $g_{1,3}$ and
$g_{2,3}$ with $h_g=-1$, and $g_{2,4}$ with $h_g=0$; the remaining groups take
the raw route. The outputs are
\begin{equation}
\begin{aligned}
y_1&=\gamma_1A_1-\gamma_1A_3+g_{1,2}A_2+g_{1,4}A_4,\\
y_2&=\gamma_2A_1-\gamma_2A_3+g_{2,2}A_2.
\end{aligned}
\label{eq:routed-example}
\end{equation}
where all products are slot-wise and the all-zero group $g_{2,4}$
contributes nothing. Evaluated term by term, the two outputs need seven \PMult
operations, four by reconstruction factors and three by raw groups.

\parh{Grouping Reconstruction Operations.}
Signed terms of the same output share one \PMult when they use the same
packed reconstruction factors and their input ciphertexts can be added directly
in CKKS, that is, at the same slot arrangement, level, and scale. \tool first
forms their signed sum and then applies the factors once.
For example, $\gamma_1A_1-\gamma_1A_3$ in \E~\ref{eq:routed-example} becomes
$\gamma_1(A_1-A_3)$. The same rewrite applies to $y_2$. Each
output needs one \PMult by its reconstruction factors instead of two, and
the total falls from seven to five.

\parh{Sharing Signed Sums.}
Different outputs can also reuse the same signed sum, even when
their reconstruction factors differ.
With $U=A_1-A_3$, the example becomes
\begin{equation}
\begin{aligned}
y_1&=\gamma_1U+g_{1,2}A_2+g_{1,4}A_4,\\
y_2&=\gamma_2U+g_{2,2}A_2.
\end{aligned}
\label{eq:shared-example}
\end{equation}
Computing $U$ once saves one subtraction.
Each output then applies its own reconstruction factor to $U$.

To find such reusable sums, \tool compares pairs of outputs and identifies
the signed terms they have in common, namely the same input ciphertexts with
the same signs, where the sign of a term is the common ternary value $h_g$ of
its group. 
A sum can also be reused with all signs reversed; for example, an output
that needs $A_3-A_1$ uses $-U$. For each candidate sum, \tool estimates the
number of \Add and \Sub operations it removes. For a sum of $m$ input ciphertexts
shared by $k$ outputs, it conservatively counts $km$ operations
for separate accumulation and $m+k$ for forming the shared sum
and accumulating it into the $k$ outputs. The estimated saving
is therefore $km-m-k$. \tool
accepts candidates with a positive estimate and, within one output, at most one
of any two overlapping candidates, since using both would add an input
ciphertext twice and change the result.

\parh{Combining the Signed and Raw Routes.}
A \PMult raises the scale of its result, which CKKS restores by rescaling
(\S~\ref{sec:background-ckks}). In an output such as $y_1$, the \PMult by the
reconstruction factors and the \PMult by each raw group produce results at the
same scale when their plaintext operands are encoded at the same scale, so
\tool adds these products first and rescales their sum once, instead of
rescaling each product separately.

\subsection{Post-Routing Polynomial Optimization}
\label{sec:method-corridor}

As observed in \S~\ref{subsec:ob4}, weight ternarization narrows
the input ranges of downstream polynomials, so a polynomial fitted to the routed model can often have a lower degree,
and thus a lower depth, than the supplied one.
We call each evaluation of a supplied polynomial in the model a \emph{site}
and write $h_s$ for the polynomial at site $s$; a site is an activation
polynomial or a polynomial inside another operator, such as the
inverse-square-root approximation in normalization. This stage fits
lower-degree candidates at each site to the inputs the site receives in the
current model, and then replaces polynomials site by site, in the order of the
depth they can save, accepting a replacement only if the model stays within a
cumulative accuracy budget relative to $f_{\mathrm{route}}$.

\parh{Fitting Candidates to Current Inputs.}

Starting from $f=f_{\mathrm{route}}$, \tool runs the current model $f$,
which includes all accepted replacements, on the calibration dataset
$\calib$ and records, for each site $s$, the set $\mathcal X_s(f)$ of values
that arrive at the input of $h_s$. A candidate for site $s$ is a polynomial
$q_s(x;\theta)$ of lower degree than $h_s$, including constants and
first-degree polynomials, with coefficient vector $\theta$. \tool fits each
candidate to $h_s$ on these inputs by weighted least squares,
\begin{equation}
\theta^*
=\underset{\theta}{\arg\min}
\sum_{x\in\mathcal X_s(f)}
w_s(x)\left\|h_s(x)-q_s(x;\theta)\right\|_2^2.
\label{eq:polynomial-fit}
\end{equation}

The weight $w_s(x)$ makes the fitting error measure the error at the
output of the operator that contains the site rather than at the output of
$h_s$ itself. For an activation polynomial the two coincide, so $w_s(x)=1$. In
normalization they differ.
For example, standard LayerNorm computes
$\mathbf y
=\boldsymbol\alpha\odot
(\mathbf x- \boldsymbol \mu)\frac{1}{\sqrt{\sigma^2}}
+\boldsymbol\beta$,
where $\mu$ and $\sigma^2$ are the mean and variance of the input
$\mathbf x$, and $\boldsymbol\alpha$ and $\boldsymbol\beta$ are the learned
weights and biases.
When replacing the supplied inverse-square-root polynomial,
its fitting error is multiplied by
$\boldsymbol\alpha\odot(\mathbf x- \boldsymbol \mu)$.
We therefore use
$w_s(x)=\|\boldsymbol\alpha\odot(\mathbf x- \boldsymbol \mu)\|_2^2$
to minimize the resulting squared error at the normalization output.

\parh{Selecting Replacements by Depth.}
\tool guides the selection by the \CMult depth of the model,
$\mathrm{depth}(f)$, the longest chain of dependent \CMult operations. \A~\ref{alg:polynomial-selection} keeps the set $\mathcal S$ of
sites not yet processed and takes from $\mathcal S$ the site
whose candidates could reduce the depth the most. This reduction follows from
the candidate degrees alone, so it is known before fitting. Sites with the same
reduction are taken in the model's execution order, and the reductions of the
remaining sites are recomputed after each step, since a replacement can change
which path through the model is the deepest.

At the chosen site $s$, \tool fits the candidates
(\E~\ref{eq:polynomial-fit}) and tries them in increasing order of the resulting model depth. For a candidate $q$, let $f'=f[s\leftarrow q]$ denote the
model after replacing $h_s$ with $q$. A candidate is checked for accuracy only
if $\mathrm{depth}(f')<\mathrm{depth}(f)$, and it is accepted if the accuracy
of $f'$ on a selection dataset $\mathcal D_{\mathrm{sel}}$
(\S~\ref{sec:implementation}) is at most $\epsilon$ below that of
$f_{\mathrm{route}}$, where $\epsilon$ is the accuracy budget. The first
accepted candidate replaces $h_s$; if none is accepted, $h_s$ stays. Because
every candidate is compared with the same reference $f_{\mathrm{route}}$, the
budget bounds the accuracy loss of all accepted replacements together rather
than of each one separately. Each site is processed once, and the final model
is $f_{\mathrm{poly}}$.

\begin{algorithm}[t]
\caption{Post-routing polynomial selection}
\label{alg:polynomial-selection}
\begin{algorithmic}[1]
\Require Routed model $f_{\mathrm{route}}$, data $\calib,\mathcal D_{\mathrm{sel}}$, budget $\epsilon$
\State $f\gets f_{\mathrm{route}}$; $\mathcal S\gets\Call{GetPolys}{f}$
\State $b\gets\Call{Accuracy}{f_{\mathrm{route}},\mathcal D_{\mathrm{sel}}}$
\While{$\mathcal S\ne\varnothing$}
  \State $s\gets\Call{NextPoly}{f,\mathcal S}$
  \State $\mathcal S\gets\mathcal S\setminus\{s\}$
  \State $\mathcal C_s\gets\Call{FitCandidates}{f,s,\calib}$
  \For{each $q\in\mathcal C_s$ in depth order}
    \State $f'\gets f[s\leftarrow q]$
    \If{$D(f')<D(f)$}
      \State $a\gets\Call{Accuracy}{f',\mathcal D_{\mathrm{sel}}}$
      \If{$a\ge b-\epsilon$}
        \State $f\gets f'$; \textbf{break}
      \EndIf
    \EndIf
  \EndFor
\EndWhile
\State \Return $f_{\mathrm{poly}}\gets f$
\end{algorithmic}
\end{algorithm}

\subsection{Public-Constant Folding and Plan Export}
\label{sec:method-folding}

A \emph{public constant} is a value the server holds in plaintext, such as
a weight, a reconstruction factor, or a polynomial coefficient. With routes and
polynomials fixed, \tool folds public constants into adjacent operators in two
directions.

\parh{Folding into the Following Polynomial.}
For a first-degree operation $t=ax+b$ with public $a$ and $b$ whose output
is used only by the polynomial $p$, \tool computes offline the coefficients of
the composed polynomial $\widetilde p(x)=p(ax+b)$, which has the same degree as
$p$, and evaluates $\widetilde p$ on $x$ directly, eliminating the separate
\PMult of $t$.

\parh{Folding into the Preceding Linear Operator.}
A multiplication by a public constant can instead be folded into the
plaintext operands of the linear operator that produces its input.
Suppose $y_1$ in \E~\ref{eq:shared-example} is used only in a
subsequent operation $z=a \cdot y_1$, where $a$ is a public constant. We can
rewrite this computation as $z=(a\gamma_1)U+(ag_{1,2})A_2+(ag_{1,4})A_4$.
\tool computes $a\gamma_1$, $ag_{1,2}$, and $ag_{1,4}$ offline and uses them in
place of the original 
operands. The rest of the computation is unchanged,
and the separate \PMult for $ay_1$ disappears.

\parh{Checking Numerically Sensitive Folds.}
Each fold is an algebraic identity, so it leaves $f_{\mathrm{poly}}$
unchanged as a function. However, folding public constants into polynomial
coefficients may produce small coefficients whose CKKS encoding errors are
amplified by large intermediate values. For these numerically sensitive folds,
\tool compares the folded and unfolded forms offline with the same
encrypted inputs and CKKS backend. The maximum absolute errors of the two
forms at the output of the folded operator, $E_{\mathrm{fold}}$ and
$E_{\mathrm{keep}}$, are measured against the same full-precision plaintext
reference.

When both computations produce valid numerical results, \tool accepts the folded
form conservatively if $E_{\mathrm{fold}}\leq\max(\eta E_{\mathrm{keep}},\tau_s)$,
where $\eta$ bounds the error growth relative to the unfolded form, and
$\tau_s$ is an absolute tolerance at site $s$ that avoids an overly strict
bound when $E_{\mathrm{keep}}$ is very small. We use $\eta=10$ and
$\tau_s=10^{-4}$. If the folded form fails the check, \tool keeps the
unfolded form.

Finally, \tool removes operations whose outputs are no longer needed and then
exports the fixed execution plan $\Plan$.
 \section{Implementation and Setup}
\label{sec:implementation}
\label{sec:eval-setup}

\parh{Implementation.}
We implement \tool's offline optimizer in Python, using
PyTorch~\citep{paszke2019pytorch} for weight training and polynomial fitting.
Encrypted inference uses a C++ executor built on Microsoft SEAL
4.1~\citep{sealcrypto} and the CKKS bootstrapping implementation from
FHE-MP-CNN~\citep{lee2022low}, as in NEXUS~\citep{zhang2025nexus} and
MOAI~\citep{zhang2026moai}. The optimizer exports a plan $\Plan$ that
specifies the weights, polynomials, packed operations, and CKKS rescaling and
bootstrapping schedules used by the executor.

\begin{figure}[t]
\centering
\includegraphics[width=\columnwidth]{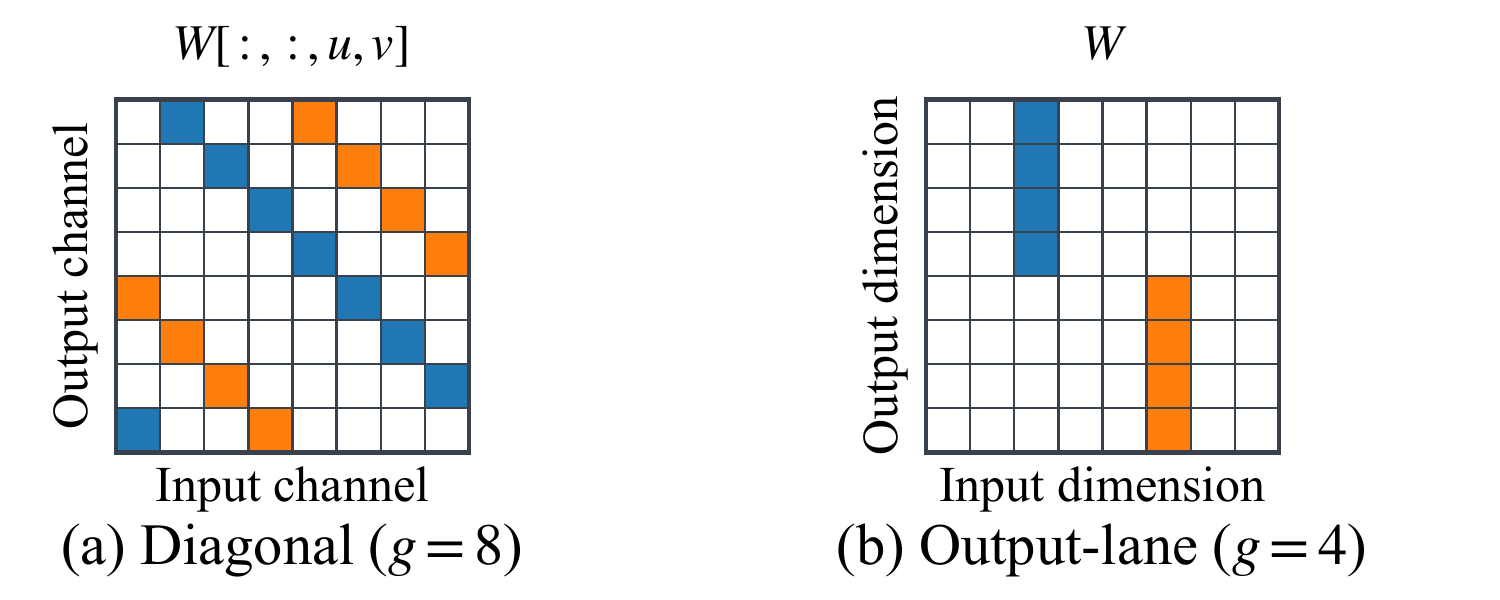}
\caption{Grouping rules of \T~\ref{tab:implementation-layouts}
for (a) VGG11 convolutions at a fixed kernel
position $(u,v)$ and (b) ViT linear layers. Each color marks one group and 
``g'' denotes the group size.}
\label{fig:packing-groups}
\end{figure}

\parh{Models and Baselines.}
We evaluate VGG11~\citep{simonyan2014very} on CIFAR-10~\citep{krizhevsky2009learning}, ViT-S/16~\citep{dosovitskiy2020image} on Tiny-ImageNet, and
BERT-base~\citep{devlin2019bert} on SST-2~\citep{socher2013recursive} with sequence length 128. VGG11 and ViT follow
ULD-Net's polynomial activation, normalization and RoPE attention designs~\citep{xie2026uldnet, su2024roformer};
BERT follows MOAI's nonlinear approximation strategy~\citep{zhang2026moai}.
For each model, 
the baseline keeps the raw weights of every layer and uses the supplied
polynomials, and \tool applies the offline optimizations of
\S~\ref{sec:method} to the same architecture. Following prior
works~\citep{xie2026uldnet, zhang2026moai}, 
both variants of VGG11 and ViT are trained from scratch with the polynomials
in place, and both variants of BERT are fine-tuned from the same
\texttt{bert-base} checkpoint with exact nonlinear functions, which are then
replaced by MOAI's polynomials.
The baseline is
trained with the task loss alone and the \tool variant with
\E~\ref{eq:group-training-objective}. Both variants of a
model use the same executor and packing layout.
\T~\ref{tab:implementation-layouts} gives the grouping rule and group size
of each layout, and \F~\ref{fig:packing-groups} illustrates the VGG11 and ViT
rules. The VGG11 layout uses blocks of eight output channels,
and the ViT layout uses blocks of four consecutive output features.

\begin{table}[t]
\centering
\caption{
Packing layouts of the evaluated models, each taken from the
cited work. The ``Grouping Rule'' states which weights form one execution group
(\S~\ref{sec:background-layout}), and the ``Size'' is the number of weights per
group.}
\label{tab:implementation-layouts}
\small
\setlength{\tabcolsep}{3pt}
\begin{tabular}{@{}p{0.14\linewidth}p{0.72\linewidth}c@{}}
\toprule
Model & Grouping rule & Size \\
\midrule
VGG11 & \emph{Diagonal.} Group one diagonal within an output-channel block at a fixed kernel
position~\citep{ebel2025orion}.
& 8 \\

\addlinespace
ViT & \emph{Output lane.} Group the weights connecting one input feature to a block of consecutive output features.~\citep{zhang2025nexus}.
& 4 \\
\addlinespace
BERT & \emph{Single weight.} Group each weight for an input
ciphertext containing one feature column~\citep{zhang2026moai}.
& 1 \\
\bottomrule
\end{tabular}
\end{table}

\parh{Optimization Settings.}
To balance accuracy and efficiency, we set the maximum protection ratio $\rho_{\mathrm P,\max}=0.20$,
so that at most 80\% of the groups in each packing pool take the signed
route after each per-epoch protection decision (\S~\ref{sec:method-shaping}).
As in ULD-Net~\citep{xie2026uldnet}, both ViT variants use
variance regularization during training to stabilize
polynomial normalization.

For the polynomial optimization of \S~\ref{sec:method-corridor}, the
calibration dataset $\calib$ is half of the CIFAR-10 test set for VGG11, half
of the Tiny-ImageNet validation set for ViT, and 1,000 SST-2 training examples
for BERT; the selection dataset $\mathcal D_{\mathrm{sel}}$ is the same half
of the dataset for VGG11 and ViT, and the SST-2 development set for BERT.
The accuracy budget is $\epsilon=0.2\%$. The plaintext accuracies in
\T~\ref{tab:eval-overall} are measured on the full CIFAR-10 test set,
Tiny-ImageNet validation set, and SST-2 development set, which include $\mathcal
D_{\mathrm{sel}}$.

\parh{Execution Environment.}
We run CKKS inference on Intel Xeon Platinum 8592+ CPUs, using 48 cores and 480\,GiB of memory per run. 
Both variants
use ring dimension
$N=2^{16}$, $2^{15}$ slots per ciphertext, and scale $2^{46}$, following MOAI's parameter settings~\citep{zhang2026moai}.
One batch holds 4 images for VGG11, 32 images for
ViT, and 256 sequences for BERT; these batch sizes fill the $2^{15}$ slots
of each ciphertext under the packing layout with padding (a BERT ciphertext holds one feature for the 128 tokens of 256 sequences).

\parh{Measurements.}
Latency is the end-to-end time of the execution plan, including layout
conversions, plaintext encoding, and bootstrapping. \#\PMult counts the
plaintext--ciphertext multiplications of the plan outside bootstrapping, and
\#Boot counts bootstrapping operations.

 \section{Evaluation}
\label{sec:evaluation}

This section evaluates \tool through the following research questions:
\begin{itemize}[leftmargin=*,itemsep=2pt,parsep=0pt,topsep=3pt]
\item \textbf{RQ1:} How does \tool compare with full-precision baselines
in end-to-end inference latency and model accuracy?
\item \textbf{RQ2:} How do packing-aware weight routing and exact
linear compilation affect inference cost and accuracy?
\item \textbf{RQ3:} How does polynomial optimization affect
polynomial approximation and \CMult depth?
\end{itemize}

\subsection{RQ1: End-to-End Performance}
\label{sec:eval-outcome}

We compare the baseline and its \tool-optimized
counterpart in accuracy, operator cost, and latency.

\begin{table}[t]
\centering
\caption{
End-to-end encrypted inference of the baseline and the \tool
variant (\S~\ref{sec:implementation}). \#\PMult (in millions)
and \#Boot count one full batch: 4 images for VGG11, 32 images for
ViT, and 256 examples for BERT. Amortized latency is the batch time
divided by the batch size. Acc. is plaintext
accuracy.
}
\label{tab:eval-overall}
\small
\setlength{\tabcolsep}{1.5pt}
\begin{tabular*}{\columnwidth}{@{\extracolsep{\fill}}llrrrrr@{}}
\toprule
Model & Variant & \shortstack{Acc.\\(\%)} & \shortstack{\#\PMult\\(M)} & \#Boot &
\shortstack{Amortized\\latency (s)} & Speedup \\
\midrule
\multirow{2}{*}{VGG11}
 & ULD-Net~\citep{xie2026uldnet} & 87.17 & 1.159 & 280 & 579.61 & $1.00{\times}$ \\
 & \tool & 86.27 & 0.482 & 120 & 243.10 & $\mathbf{2.38{\times}}$ \\
\midrule
\multirow{2}{*}{ViT}
 & ULD-Net~\citep{xie2026uldnet} & 60.02 & 5.542 & 4,608 & 628.37 & $1.00{\times}$ \\
 & \tool & 59.11 & 2.582 & 2,016 & 374.44 & $\mathbf{1.68{\times}}$ \\
\midrule
\multirow{2}{*}{BERT}
 & MOAI~\citep{zhang2026moai} & 91.06 & 86.538 & 62,340 & 847.92 & $1.00{\times}$ \\
 & \tool & 90.37 & 17.752 & 39,252 & 459.95 & $\mathbf{1.84{\times}}$ \\
\bottomrule
\end{tabular*}
\end{table}

\parh{Overall Efficiency and Accuracy.}
In \T~\ref{tab:eval-overall}, \tool accelerates encrypted
inference by $2.38{\times}$ on VGG11, $1.68{\times}$ on ViT, and
$1.84{\times}$ on BERT relative to each baseline, while the plaintext accuracy loss is less than 1\% for all three models.
It also reduces
\#\PMult by 53.4--79.5\% and \#Boot by 37.0--57.1\%.

\parh{Sources of Execution Time.}
In \T~\ref{tab:eval-operator-time}, we compare execution time of
five major operator classes. \tool accelerates plaintext weight linear
operators by $2.85{\times}$, $1.75{\times}$, and $3.98{\times}$ in
VGG11, ViT, and BERT, respectively. 
These speedups come from the signed
route, which replaces the \PMult of a pure group by \Add and \Sub
(\S~\ref{sec:method-shaping}), and exact compilation, which removes
reconstruction \PMult and \Add/\Sub (\S~\ref{sec:method-reuse}).

Activation, normalization, and softmax computations also
take less time with \tool's polynomial optimization.
For example, the polynomial optimization replaces the Softmax exponential of
some attention heads by a constant (\S~\ref{sec:method-corridor}) in BERT,
which contributes to the $2.86{\times}$ speedup in ciphertext matrix
multiplication. ViT's RoPE attention has no Softmax and thus no polynomial for \tool to replace, so its ciphertext products are the same in both variants.

By reducing \CMult depth, \tool also consumes fewer ciphertext
levels and needs $2.33{\times}$, $2.29{\times}$, and $1.59{\times}$ fewer
bootstrapping operations (\#Boot in \T~\ref{tab:eval-overall}). Their total
bootstrapping time is $2.04{\times}$, $2.15{\times}$, and $1.71{\times}$ lower
than the baseline's for VGG11, ViT, and BERT, respectively. Bootstrapping processes ciphertexts in parallel, so time savings
also depend on the parallelism available at each refresh position.

\begin{table}[t]
\centering
\caption{Amortized time of major operators (s/input),
using the batch sizes in \T~\ref{tab:eval-overall}. ``Plaintext
Weight'' covers convolutions and linear layers; ``Act'' and ``Norm'' denote
activation and normalization. ``Ciphertext MatMul'' uses two encrypted
operands. 
Times include the plaintext encoding and layout conversions of
each operator.
N.A. denotes not applicable.}
\label{tab:eval-operator-time}
\small
\setlength{\tabcolsep}{1.5pt}
\begin{tabular*}{\columnwidth}{@{\extracolsep{\fill}}llrrrrr@{}}
\toprule
Model & Variant & \shortstack{Plaintext\\Weight} & Act & Norm & Softmax &
\shortstack{Ciphertext\\MatMul} \\
\midrule
\multirow{3}{*}{VGG11}
 & ULD-Net~\citep{xie2026uldnet} & 332.95 & 1.15 & 14.23 & N.A. & N.A. \\
 & \tool & 116.78 & 0.43 & 11.73 & N.A. & N.A. \\
 & \textbf{Speedup} & $\mathbf{2.85{\times}}$ & $\mathbf{2.67{\times}}$ & $\mathbf{1.21{\times}}$ & N.A. & N.A. \\
\midrule
\multirow{3}{*}{ViT}
 & ULD-Net~\citep{xie2026uldnet} & 114.17 & 1.46 & 17.21 & N.A. & 121.13 \\
 & \tool & 65.06 & 0.83 & 9.73 & N.A. & 124.24 \\
 & \textbf{Speedup} & $\mathbf{1.75{\times}}$ & $\mathbf{1.77{\times}}$ & $\mathbf{1.77{\times}}$ & N.A. & $\mathbf{0.97{\times}}$ \\
\midrule
\multirow{3}{*}{BERT}
 & MOAI~\citep{zhang2026moai} & 87.36 & 5.51 & 15.11 & 1.62 & 24.42 \\
 & \tool & 21.97 & 4.33 & 7.22 & 0.81 & 8.54 \\
 & \textbf{Speedup} & $\mathbf{3.98{\times}}$ & $\mathbf{1.27{\times}}$ & $\mathbf{2.09{\times}}$ & $\mathbf{1.99{\times}}$ & $\mathbf{2.86{\times}}$ \\
\bottomrule
\end{tabular*}
\end{table}

\begin{figure}[t]
\centering
\includegraphics[width=0.9\columnwidth]{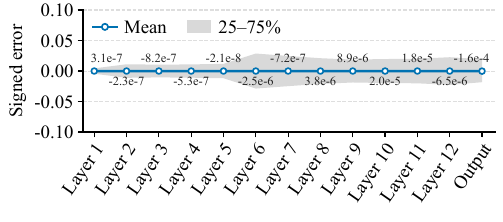}
\Description{Signed errors at twelve BERT encoder block outputs and final
logits. Each mean is numerically labeled. The gray 25th--75th percentile
band widens at Layer 6 to about minus 0.029 to plus 0.029. At Output,
the mean is minus 0.0001624 and the band is about minus 0.0183 to plus
0.0192. Statistics cover all 256 inputs and exclude padding.}
\caption{Signed error (decrypted minus plaintext) for \tool BERT.
Markers and labels show the mean; gray shading spans the 25th--75th
percentiles. 
Statistics cover all 256 inputs at the encoder block outputs (Layer 1--12) and final logits (Output).
}
\label{fig:eval-signed-error}
\end{figure}

\parh{Encrypted versus Plaintext Outputs.}
At the final output, the RMSE between the decrypted and the
plaintext logits is $4.26{\times}10^{-4}$, $8.77{\times}10^{-3}$, and
$3.74{\times}10^{-2}$ for VGG11, ViT, and BERT, respectively, and the
encrypted prediction equals the plaintext prediction for every input of the
measured batches. \F~\ref{fig:eval-signed-error} traces the
signed errors through the layers of BERT, the model with the largest RMSE, and
shows that the mean stays within $2.0{\times}10^{-5}$ of zero across all 12
blocks.

\subsection{RQ2: Effectiveness of Weight Optimization}
\label{sec:eval-weight-causality}

We first examine whether weight routing creates pure groups and protecting sensitive weights preserves accuracy. We then keep
weights and routes fixed in representative linear layers to measure the cost
reduction from exact compilation.

\parh{Packing-Aware Weight Routing.}
We compare scalar ternary quantization, which ternarizes each weight on its
own (\S~\ref{sec:motivation}), with \tool's weight routing on VGG11. Both
models use the polynomials and the packing layout of
\S~\ref{sec:implementation}. After training, we compare accuracy, fraction of pure groups, and weight \PMult count, the number of \PMult
operations by raw weights and reconstruction factors.

\begin{table}[t]
\centering
\caption{
Weight routing on VGG11. Pure is the percentage of pure groups in
the eight convolutional layers. Weight \PMult covers these layers and the
classifier, and the reduction is relative to the baseline.
}
\label{tab:eval-shaping}
\small
\setlength{\tabcolsep}{3pt}
\begin{tabular*}{\columnwidth}{@{\extracolsep{\fill}}lrrrr@{}}
\toprule
Training & \shortstack{Acc.\\(\%)} & \shortstack{Pure\\(\%)} &
\shortstack{Weight\\\PMult} & \shortstack{Reduction\\vs. Full-precision (\%)} \\
\midrule
Scalar ternary & 84.64 & 0.05 & 1,152,665 & 0.02 \\
$f_{\mathrm{route}}$ & \textbf{86.87} & \textbf{58.88} & \textbf{477,347} & \textbf{58.59} \\
\bottomrule
\end{tabular*}
\end{table}

\T~\ref{tab:eval-shaping} shows that scalar ternary quantization leaves
only 0.05\% of the packed groups pure. The mixed groups still
require vector \PMult, so the weight \PMult count
remains close to that of the full-precision model. In contrast, \tool produces 58.88\% pure groups and reduces weight \PMult counts by 58.59\% relative to the full-precision model. Accuracy also improves by 2.23\% over
scalar ternary quantization. 

\parh{Protecting sensitive weights.}
BERT's single-weight groups make every ternary candidate structurally
pure. We fine-tune
two models on SST-2 for three epochs, with protection (\S~\ref{sec:method-shaping}) enabled or disabled.
With protection enabled, 20\% of the encoder weights remain
in full precision.

For each final model, we measure accuracy in three settings (\T~\ref{tab:eval-protection}).
\emph{Non-Linear} uses the original nonlinear operators.
For \emph{Direct Transfer}, we calibrate the encoder approximations
on a separate full-precision BERT model fine-tuned on SST-2, then apply
them unchanged to the models with protection enabled and disabled.
\emph{Refitted} refits the approximations, at their supplied degrees, to
each model's own inputs, with the weights unchanged.

\begin{table}[t]
\centering
\caption{Sensitive-weight protection on BERT-base/SST-2.
$\dagger$ denotes non-finite outputs on all samples.}
\label{tab:eval-protection}
\small
\setlength{\tabcolsep}{4pt}
\begin{tabular*}{\columnwidth}{@{\extracolsep{\fill}}lrrr@{}}
\toprule
Protection & Non-Linear & Direct Transfer & Refitted \\
\midrule
Disabled & 88.07\% & 0.00\%$^{\dagger}$ & 88.30\% \\
Enabled & \textbf{91.51\%} & \textbf{90.48\%} & \textbf{91.51\%} \\
\bottomrule
\end{tabular*}
\end{table}

\T~\ref{tab:eval-protection} shows that 
protection improves the Non-Linear accuracy
from 88.07\% to 91.51\%.
With directly transferred approximations, the protected model
achieves 90.48\% accuracy, whereas the unprotected model produces
non-finite outputs for all samples in the SST-2 development dataset.

Refitting brings the accuracy of the unprotected model back to 88.30\%, close to its
Non-Linear accuracy, while the protected model reaches 91.51\%.
The protected model therefore retains a 3.21\%
advantage even when each model uses approximations fitted
to its own inputs.

\parh{Exact Compilation.}
For each model, we select a layer with the largest
number of weights, namely a $3{\times}3$, 512-channel convolution
in VGG11, a $384{\to}1536$ feed-forward projection in ViT,
and a $768{\to}3072$ feed-forward projection in BERT.
Within each layer, weights and routes remain fixed across the four
compilations. The first compilation applies the reconstruction factors
separately to each signed term. We then enable, in sequence, the grouping of
reconstruction operations, the sharing of signed sums, and the combining of the
signed and raw routes (\S~\ref{sec:method-reuse}). Within each model, all
compilations use the same input and output CKKS levels. Time includes the
encrypted computation and the plaintext encoding.

\begin{table}[t]
\centering
\caption{Exact compilation of representative routed linear operators.
Time is the mean of three runs after one warm-up per variant.
Weight \PMult includes raw products and reconstruction, and 
\Add/\Sub counts ciphertext additions and subtractions.}
\label{tab:eval-linear-compilation}
\small
\setlength{\tabcolsep}{2pt}
\begin{tabular*}{\columnwidth}{@{\extracolsep{\fill}}llrrrr@{}}
\toprule
Model & Compilation & \shortstack{Weight\\\PMult} &
\shortstack{CKKS\\\Add/\Sub} & Rescale & \shortstack{Time\\(s)} \\
\midrule
\multirow{4}{*}{VGG11}
 & Per-contribution & 238,733 & 240,333 & 960 & 30.35 \\
 & + Grouping & 141,268 & 240,333 & 960 & 19.40 \\
 & + Sharing & 141,268 & 219,609 & 960 & 18.92 \\
 & + Combining & 141,268 & 219,609 & 896 & 18.99 \\
\midrule
\multirow{4}{*}{ViT}
 & Per-contribution & 131,675 & 131,291 & 767 & 12.95 \\
 & + Grouping & 80,762 & 131,291 & 767 & 8.80 \\
 & + Sharing & 80,762 & 125,223 & 767 & 8.76 \\
 & + Combining & 80,762 & 125,223 & 384 & 8.71 \\
\midrule
\multirow{4}{*}{BERT}
 & Per-contribution & 1,838,483 & 1,835,411 & 6,144 & 64.85 \\
 & + Grouping & 1,062,992 & 1,835,411 & 6,144 & 46.78 \\
 & + Sharing & 1,062,992 & 1,699,696 & 6,144 & 45.19 \\
 & + Combining & 1,062,992 & 1,699,696 & 3,072 & 45.10 \\
\bottomrule
\end{tabular*}
\end{table}

As shown in \T~\ref{tab:eval-linear-compilation}, exact compilation discussed in \S~\ref{sec:method-reuse} gives speedups of $1.60{\times}$, $1.49{\times}$,
and $1.44{\times}$ on the VGG11, ViT, and BERT layers, respectively. 
Most of the speedup comes from grouping, which applies each reconstruction
factor once to a signed sum and reduces weight \PMult counts by 38.67--42.18\%.
Sharing signed sums further reduces \Add/\Sub counts by 4.62--8.62\% without
changing \PMult counts, and combining the routes roughly halves the number of
rescaling operations in ViT and BERT.

We also compare the decrypted outputs with independent FP64
computations. The maximum absolute error across the tested
variants is $5.1{\times}10^{-9}$.

\subsection{RQ3:Effectiveness of Polynomial Optimization}
\label{sec:eval-polynomial-causality}

\parh{Input Range Contraction.}
We compare the inputs of the supplied polynomials in the baseline and in
$f_{\mathrm{route}}$, and define the input width of a site as the difference
between the 1st and 99th percentiles of its input values. Contraction is
$1-w_{\mathrm{route}}/w_{\mathrm{base}}$, where $w_{\mathrm{route}}$ and
$w_{\mathrm{base}}$ are the widths in $f_{\mathrm{route}}$ and in the
baseline. We use the width rather than the radius of \S~\ref{subsec:ob4}
because the inputs of the normalization and Softmax polynomials are not
centered at zero.
At an activation site, we measure the width per channel and take the median
of the per-channel ratios $w_{\mathrm{route}}/w_{\mathrm{base}}$ as the ratio
of the site. At a normalization site, the polynomial takes the scaled variance
as input, so the width is measured over these variance values.

\begin{figure}[t]
\centering
\begin{subfigure}[t]{0.46\columnwidth}
\centering
\includegraphics[width=\linewidth,trim=0 12bp 0 0,clip]{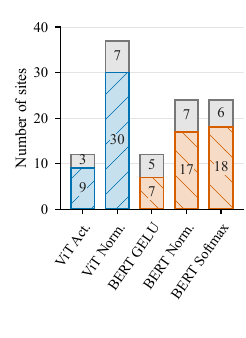}
\caption{Site counts}
\label{fig:eval-contraction-counts}
\end{subfigure}\hfill
\begin{subfigure}[t]{0.46\columnwidth}
\centering
\includegraphics[width=\linewidth,trim=0 12bp 0 0,clip]{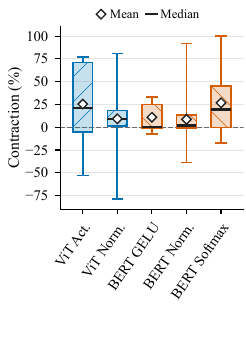}
\caption{Contraction magnitude}
\label{fig:eval-contraction-magnitude}
\end{subfigure}
\caption{Input contraction across polynomial positions.
Colored and gray bar segments count contracted and the other positions,
respectively. Boxes span the 25th--75th percentiles; lines mark
medians, diamonds mark means, and whiskers span the minimum and maximum.}
\Description{Two subplots grouped by model and operator family. Stacked
bars show contracted and remaining site counts: 9 and 3 for ViT
activation, 30 and 7 for ViT normalization, 7 and 5 for BERT GELU,
17 and 7 for BERT normalization, and 18 and 6 for BERT Softmax.
Boxplots show the full contraction distributions, with means and medians.
Both models use light fills and distinct diagonal hatch directions.}
\label{fig:eval-contraction}
\end{figure}

As shown in \F~\ref{fig:eval-contraction-counts}, the $f_{\mathrm{route}}$ has
narrower input ranges at 39 of 49 measured sites in ViT and
42 of 60 in BERT. More than half of the sites contract in every operator
family. The magnitude varies across sites and operator families as shown in
\F~\ref{fig:eval-contraction-magnitude}, with mean contractions of 25.42\% for
ViT activations and 26.80\% for BERT Softmax, which offer opportunities for
lower-degree approximation.

\parh{Low-Degree Approximation.}
We next compare lower-degree approximations at the sites
with the largest input contraction for activations, normalization,
and Softmax exponentials in each model.
At each site, we fit candidates of the same degree for the baseline and
for $f_{\mathrm{route}}$. Each candidate approximates the polynomial already
used by its model, and we measure the mean squared error (MSE) between the
candidate and that polynomial over the inputs recorded at the site.

\begin{figure}[t]
\centering
\includegraphics[width=0.8\columnwidth,trim=0 5bp 0 8bp,clip]{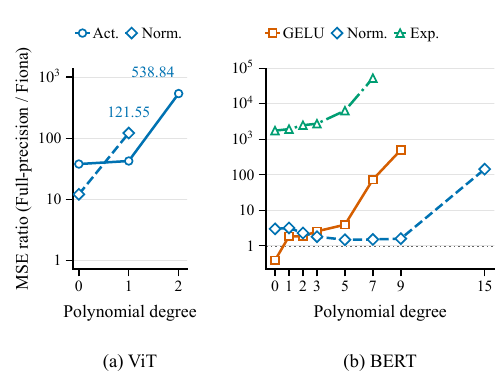}
\caption{
MSE of same-degree fits to the supplied polynomial, each measured
on the inputs its model records at the site. The ratio is the baseline MSE
divided by the $f_{\mathrm{route}}$ MSE.
}
\Description{Two panels plot MSE ratios against polynomial degree:
ViT activation and normalization on the left; BERT GELU, normalization,
and Softmax exponential on the right.
The panels use different logarithmic ranges and show 26 comparisons.}
\label{fig:eval-polynomial-sites}
\end{figure}

\begin{table}[t]
\centering
\caption{
Accuracy and polynomial \CMult depth before and after polynomial
optimization. Act., Norm., and Softmax give the \CMult depth of the polynomials
of each family summed over its sites, Total sums the families, and $\Delta$
Depth is the relative reduction of Total. Accuracies are on the full
Tiny-ImageNet validation set and SST-2 development set.
}
\label{tab:eval-polynomial}
\small
\setlength{\tabcolsep}{2pt}
\begin{subtable}{0.9\columnwidth}
\centering
\caption{ViT on Tiny-ImageNet}
\label{tab:eval-polynomial-vit}
\begin{tabular*}{0.9\linewidth}{@{\extracolsep{\fill}}lrrrrr@{}}
\toprule
Model & Acc. & Act. & Norm. & Total & $\Delta$ Depth \\
\midrule
$f_{\mathrm{route}}$ & 59.51\% & 24 & 111 & 135 & -- \\
$f_{\mathrm{poly}}$ & 59.11\% & 14 & 48 & 62 & 54.07\% \\
\bottomrule
\end{tabular*}
\end{subtable}

\medskip
\begin{subtable}{0.9\columnwidth}
\centering
\caption{BERT on SST-2}
\label{tab:eval-polynomial-bert}
\begin{tabular*}{0.9\linewidth}{@{\extracolsep{\fill}}lrrrrrr@{}}
\toprule
Model & Acc. & Act. & Norm. & Softmax & Total & $\Delta$ Depth \\
\midrule
$f_{\mathrm{route}}$ & 90.48\% & 73 & 376 & 300 & 749 & -- \\
$f_{\mathrm{poly}}$ & 90.37\% & 61 & 147 & 168 & 376 & 49.80\% \\
\bottomrule
\end{tabular*}
\end{subtable}
\end{table}

As shown in \F~\ref{fig:eval-polynomial-sites}, the fits on
$f_{\mathrm{route}}$ have $538.84{\times}$ and $121.55{\times}$ lower MSE than
the fits on the baseline for the quadratic activation fit and the first-degree
normalization fit in ViT, respectively. For BERT, the first-degree GELU and
normalization fits have $1.85{\times}$ and $3.17{\times}$ lower MSE. A lower
degree can also achieve a smaller error. At the ViT activation site, the
first-degree fit on $f_{\mathrm{route}}$ has an MSE of $7.20\times10^{-4}$,
below the $9.13\times10^{-3}$ of the baseline's quadratic fit.

\parh{Reduction of \CMult depth.}
We compare $f_{\mathrm{route}}$, which retains the supplied
polynomials, with $f_{\mathrm{poly}}$, which uses the polynomials replaced by \tool.
Both use the same weights optimized by \tool\ for each model, and the initially supplied polynomials follow
ULD-Net for ViT and MOAI for BERT~\citep{xie2026uldnet,zhang2026moai}.

\T~\ref{tab:eval-polynomial} shows that polynomial optimization reduces total polynomial \CMult
depth from 135 to 62 in ViT and from 749 to 376 in BERT, reductions of 54.07\%
and 49.80\%, respectively. Polynomials are replaced at 34 sites in ViT and 31 sites in BERT. Normalization accounts for most of the depth reduction in
both models, and BERT also benefits from lower-degree Softmax approximations,
whose depth contribution decreases from 300 to 168.
In addition, the same procedure under the same budget reduces the total
depth of the baselines only to 69 in ViT and 408 in BERT, so the routed
weights allow 10.14\% and 7.84\% lower depth.

\FloatBarrier
 \section{Related Work}
\label{sec:related}

\parh{FHE compilation and execution.}
EVA, HECO, and HEIR provide languages and intermediate
representations for encrypted
computation~\citep{dathathri2020eva,viand2023heco,bian2024heir}.
Fhelipe and HELayers automate tensor
packing~\citep{krastev2024fhelipe,aharoni2023helayers}.
Orion optimizes convolution mappings and level management~\citep{ebel2025orion}.
NEXUS improves packed matrix multiplication~\citep{zhang2025nexus},
while MOAI combines column and diagonal packing to reduce rotations
and avoid layout conversions~\citep{zhang2026moai}.
\tool uses the selected packing layout to guide weight optimization,
then compiles the resulting hybrid representation with common
reconstruction and shared signed sums over compatible
input ciphertexts.

\parh{Weight structure for FHE inference.}
SpENCNN combines packing with sub-block pruning~\citep{ran2023spencnn}.
PrivCirNet uses block-circulant weights with a matching
encoding~\citep{xu2024privcirnet}.
REDsec executes discretized networks with TFHE~\citep{folkerts2023redsec},
while ENSI uses a specific column encoding for ternary BitLinear
matrices~\citep{he2025ensi}.
\tool uses task sensitivity to shape weights within fixed packing groups
toward a common ternary state. The resulting hybrid operators reduce
multiplications for nonzero signed
contributions while retaining raw weights in mixed and protected groups.

\parh{Polynomial optimization for FHE inference.}
AutoFHE jointly selects polynomial degrees and bootstrap placement
while adapting network weights in full precision~\citep{ao2024autofhe}.
SLOTHE selectively approximates components of non-arithmetic
functions~\citep{nam2025slothe}, while ULD-Net trains networks with
ultra-low-degree polynomial operators~\citep{xie2026uldnet}.
With weights and routes fixed, \tool fits lower-degree candidates to
the supplied functions on inputs from the current network.
It selects replacements that reduce network depth within a cumulative
accuracy budget, exploiting approximation opportunities created by
weight routing.
 \section{Conclusion}
\label{sec:conclusion}

We present \tool, an offline optimizer that reduces FHE inference cost through packing-aware weight optimization and lower-degree polynomials. Experiments on CNN and Transformer models demonstrate speedups with little loss in plaintext accuracy.

\end{document}